\newif\ifJOURNAL	
\JOURNALfalse
\newif\ifarXiv		
\arXivfalse
\newif\ifWP		
\WPfalse
\newif\ifFULL		
\FULLfalse

\arXivtrue

\newif\ifTR		
\TRfalse
\ifarXiv\TRtrue\fi
\ifWP\TRtrue\fi

\newif\ifnotJOURNAL	
\notJOURNALtrue
\ifJOURNAL\notJOURNALfalse\fi

\ifJOURNAL
\documentclass[toc]{elsart}
\usepackage{amsmath,amsfonts,latexsym,graphicx}
\newcommand{\Extra}[1]{}
\fi

\ifarXiv
\documentclass{article}
\usepackage{amsmath,amsfonts,latexsym,graphicx}
\newcommand{\Extra}[1]{}
\fi

\ifWP
\documentclass[toc]{gtarticle}
\usepackage{amsmath,amsfonts,latexsym,epsfig,graphicx}
\renewcommand{\Extra}[1]{#1}
\fi

\allowdisplaybreaks[3]

\newcommand{\Vladimir}{Vladimir }
\newcommand{\DOT}{.}

\ifJOURNAL
  \newcommand{\GTPVII}{vovk/shafer:2005RSS}
  \newcommand{\GTPVIII}{vovk/etal:2005AIStats-local}
  
  \newcommand{\GTPXIII}{vovk:2005ALT-GTP13-local}
  \newcommand{\GTPXIV}{vovk:2005ALT-GTP14}
  \newcommand{\GTPXVII}{vovk:2006COLT-invited-full}
\fi
\ifarXiv
  \newcommand{\GTPVII}{vovk/shafer:2005RSS}
  \newcommand{\GTPVIII}{GTP8arXiv}
  
  \newcommand{\GTPXIII}{vovk:GTP13-arXiv-local}
  \newcommand{\GTPXIV}{vovk:GTP14-arXiv-local}
  \newcommand{\GTPXVII}{GTP17arXiv}
\fi
\ifWP
  \newcommand{\GTPVII}{GTP7}
  \newcommand{\GTPVIII}{GTP8}
  
  \newcommand{\GTPXIII}{vovk:GTP13-WP-local}
  \newcommand{\GTPXIV}{GTP14}
  \newcommand{\GTPXVII}{GTP17}
\fi

\ifJOURNAL
  \newcommand{\Fig}{Fig.~}
\fi
\ifnotJOURNAL
  \newcommand{\Fig}{Figure }
\fi

\newcommand{\bbbr}{\mathbb{R}}		
\newtheorem{lemma}{Lemma}
\newtheorem{corollary}{Corollary}
\newtheorem{theorem}{Theorem}
\newenvironment{proof}
  {\trivlist\item[\hskip\labelsep\textbf{Proof}]}
  {\endtrivlist}

\newcommand{\st}{\mathop{|}}

\newcommand{\D}{\,\mathrm{d}}

\newcommand{\III}{\mathbb{I}}		
\newcommand{\K}{\mathcal{K}}		
\newcommand{\KKK}{\mathbf{K}}		
\newcommand{\FFF}{\mathcal{F}}		
\newcommand{\HHH}{\mathcal{H}}		
\newcommand{\CC}{\mathbf{c}}		

\newcommand{\sign}{\mathop{\mathrm{sign}}\nolimits}
\newcommand{\minsq}{\mathop{\mathrm{min}^2}\nolimits}
\newcommand{\FS}{\mathrm{FS}}

\newenvironment{Proof}[1]
  {\trivlist\item[\hskip\labelsep\textbf{Proof #1}]}
  {\endtrivlist}
\newcommand{\boxforqed}{\rule{.3em}{1.5ex}}
\newcommand{\qedtext}{\unskip\nobreak\hfil
  \penalty50\hskip1em\null\nobreak\hfil\boxforqed
  \parfillskip=0pt\finalhyphendemerits=0\endgraf}
\newcommand{\qedmath}{\tag*{\boxforqed}}
\newenvironment{remark*}
  {\trivlist\item[\hskip\labelsep{\bfseries Remark}]\relax}
  {\endtrivlist}

\newlength{\IndentI}
\newlength{\IndentII}
\newlength{\IndentIII}
\newlength{\WidthI}
\newlength{\WidthII}
\newlength{\WidthIII}
\ifJOURNAL
\begin{document}
\begin{frontmatter}
\title{Non-asymptotic calibration and resolution}

\author{Vladimir Vovk}

\ead{vovk@cs.rhul.ac.uk}

\address{Computer Learning Research Centre,
Department of Computer Science\\
Royal Holloway, University of London,
Egham, Surrey TW20 0EX, England}
\fi

\ifarXiv
\title{Non-asymptotic calibration and resolution}

\author{Vladimir Vovk\\
\texttt{vovk\textrm{@}cs.rhul.ac.uk}\\
\texttt{http://vovk.net}}
\fi

\ifWP
\title{Non-asymptotic calibration and resolution}

\author{Vladimir Vovk}

\newcommand{\No}{13}
\twodatestrue
\newcommand{\firstposted}{November 6, 2004}
\fi

\ifnotJOURNAL
\begin{document}
\maketitle
\fi

\begin{abstract}
  We analyze a new algorithm for probability forecasting of binary observations
  on the basis of the available data,
  without making any assumptions about the way the observations are generated.
  The algorithm is shown to be well calibrated and to have good resolution
  for long enough sequences of observations
  and for a suitable choice of its parameter,
  a kernel on the Cartesian product of the forecast space $[0,1]$ and the data space.
  Our main results are non-asymptotic:
  we establish explicit inequalities,
  shown to be tight,
  for the performance of the algorithm.
\end{abstract}

\ifJOURNAL
  \end{frontmatter}
  \tableofcontents
\fi

\section{Introduction}\label{sec:introduction}

We consider the problem of forecasting a new observation
from the available data,
which may include, e.g., all or some of the previous observations
and the values of some explanatory variables.
\ifFULL
Theoretical results about properties of forecasting algorithms
are usually derived for one of two protocols:
batch (the training set is fixed)
or on-line
(new data and observations are constantly added to the training set
after the observations are disclosed).
In this paper we use the on-line protocol
(defined formally in \S\ref{sec:K29}).
The set of allowed forecasts usually either coincides with the set of possible observations
(categorical forecasting)
or is the set of all probability distributions on the set of possible observations
(probability forecasting);
this paper deals with probability forecasting.
For simplicity,
we only consider the binary case,
with two possible observations, $0$ or $1$.

\fi
To make the process of forecasting more vivid,
we imagine that the data and observations are chosen by a player called Reality
and the forecasts are made by a player called Forecaster.
To establish properties of forecasting algorithms,
the traditional theory of machine learning
makes some assumptions about the way Reality generates the observations;
e.g., statistical learning theory \cite{vapnik:1998}
assumes that the data and observations are generated independently
from the same probability distribution.
A more recent approach,
prediction with expert advice
(see, e.g., \cite{cesabianchi/lugosi:2006}),
replaces the assumptions about Reality
by a comparison class of prediction strategies;
a typical result of this theory asserts that Forecaster
can perform almost as well as the best strategies in the comparison class.
This paper further explores a third possibility,
suggested in \cite{foster/vohra:1998},
which requires neither assumptions about Reality
nor a comparison class of Forecaster's strategies.
It is shown in \cite{foster/vohra:1998} that there exists a forecasting strategy
which is automatically well calibrated;
this result has been further developed
in, e.g., \cite{lehrer:2001,sandroni/etal:2003}.
Almost all known calibration results,
however, are asymptotic
(see \cite{schervish:1985} and \cite{schervish:1985dawid} for a critique
of the standard asymptotic notion of calibration);
a non-asymptotic result about calibration
is given in \cite{sandroni:2003}, Proposition 2,
but even this result involves unspecified constants and randomization.
The main results of this paper (Theorems~\ref{thm:K29} and~\ref{thm:N91})
establish simple explicit inequalities
characterizing calibration and resolution
of our deterministic forecasting algorithm.

Next we briefly describe the main features
of our proof techniques
and their connections with the literature.
The proofs rely on the game-theoretic approach to probability
suggested in \cite{shafer/vovk:2001}.
The forecasting protocol is complemented by another player, Skeptic,
whose role is to gamble at the odds given by Forecaster's probabilities.
It can be said that our approach to forecasting is Skeptic-based,
whereas the traditional approach is Reality-based
and prediction with expert advice is Forecaster-based.
The two most popular formalizations of gambling
are subsequence selection rules (going back to von Mises's collectives)
and martingales
(going back to Ville's critique \cite{ville:1939} of von Mises's collectives
and described in detail in \cite{shafer/vovk:2001}).
The pioneering paper \cite{foster/vohra:1998}
on what we call the Skeptic-based approach,
as well as the numerous papers developing it,
used von Mises's notion of gambling;
\cite{\GTPVII} appears to be the first paper in this direction
to use Ville's notion of gambling.
Another ingredient of this paper's approach,
considering Skeptic's continuous strategies
and thus avoiding randomization by Forecaster
(which was the standard feature of the previous work)
goes back to \cite{levin:1976uniformlatin} and is also described in \cite{kakade/foster:2004};
however, I learned it from Akimichi Takemura in June 2004
(whose observation was prompted by Glenn Shafer's talk
at the University of Tokyo).

It should be noted that, although our approach was inspired
by \cite{foster/vohra:1998} and papers further developing \cite{foster/vohra:1998},
precise statements of our results and our proof techniques are completely different:
they are more in the spirit of Levin's \cite{levin:1976uniformlatin} result
about the existence of neutral measures
(see \cite{\GTPXVII} for details).

\ifJOURNAL
This is the journal version of the paper \cite{\GTPXIII}.
It differs from the conference version in the following respects:
our inequalities are now shown to be tight;
the result about what we call Fermi--Sobolev spaces
is extended to a wide class of reproducing kernel Hilbert spaces
(and its proof has been simplified;
the tedious analysis of Fourier series expansions is no longer needed);
accordingly, most of the information about Fermi--Sobolev spaces
has been removed;
out of the two ``algorithms of large numbers'',
K29 and K29${}^*$,
we now concentrate on the K29${}^*$ algorithm.
The K29 algorithm appears to be less important in the context of this paper
but still has some virtues:
it is simpler,
is easier to extend to other prediction problems,
and is applicable to a slightly wider class of kernels.
\fi

\ifarXiv
This version (version 4) of this technical report
differs from the previous one in that it incorporates the changes
made in response to the comments of the reviewers of its journal version
(to be published in \emph{Theoretical Computer Science}).
\fi

\section{The algorithms of large numbers}
\label{sec:K29}

In this section we describe our learning protocol
and the general forecasting algorithm studied in this paper.
The protocol is:

\bigskip

\parshape=5
\IndentI  \WidthI
\IndentII \WidthII
\IndentII \WidthII
\IndentII \WidthII
\IndentI  \WidthI
\noindent
FOR $n=1,2,\ldots$:\\
  Reality I announces $x_n\in\mathbf{X}$.\\
  Forecaster announces $p_n\in[0,1]$.\\
  Reality II announces $y_n\in\{0,1\}$.\\
END FOR.

\bigskip

\noindent
On each round,
Reality chooses the datum $x_n$,
then Forecaster gives his forecast $p_n$ for the next observation,
and finally Reality discloses the actual observation $y_n\in\{0,1\}$.
Reality chooses $x_n$ from a \emph{data space} $\mathbf{X}$
and $y_n$ from the two-element set $\{0,1\}$;
intuitively, Forecaster's move $p_n$
is the probability he attaches to the event $y_n=1$.
\emph{Forecasting algorithm} is Forecaster's strategy
in this protocol.
For convenience in stating the results of \S\ref{sec:optimality},
we split Reality into two players,
Reality I and Reality II.

Our learning protocol is a perfect-information protocol;
in particular,
Reality may take into account the forecast $p_n$
when deciding on her move $y_n$.
(This feature is unusual for probability forecasting
but it extends the domain of applicability of our results
and we have it for free.)

Next we describe the general forecasting algorithm
that we study in this paper
(it was derived informally in \cite{\GTPVIII}).
A function $\KKK:Z^2\to\bbbr$,
where $Z$ is an arbitrary set and $\bbbr$ is the set of real numbers,
is a \emph{kernel on $Z$} if it is symmetric
($\KKK(z,z')=\KKK(z',z)$ for all $z,z'\in Z$)
and positive definite
($\sum_{i=1}^m\sum_{j=1}^m \lambda_i\lambda_j \KKK(z_i,z_j)\ge0$
for all $(\lambda_1,\ldots,\lambda_m)\in\bbbr^m$
and all $(z_1,\ldots,z_m)\in Z^m$).
The usual interpretation of a kernel $\KKK(z,z')$
is as a measure of similarity between $z$ and $z'$
(see, e.g., \cite{scholkopf/smola:2002}, \S1.1).
Our algorithm has one parameter,
which is a kernel on the Cartesian product $[0,1]\times\mathbf{X}$.
The most straightforward way of constructing such kernels
from kernels on $[0,1]$ and kernels on $\mathbf{X}$
is the operation of tensor product.
(See, e.g., \cite{aronszajn:1950,vapnik:1998,scholkopf/smola:2002}.)
Let us say that a kernel $\KKK$ on $[0,1]\times\mathbf{X}$
is \emph{forecast-continuous} if the function $\KKK((p,x),(p',x'))$,
where $p,p'\in[0,1]$ and $x,x'\in\mathbf{X}$,
is continuous in $(p,p')$
for any fixed $(x,x')\in\mathbf{X}^2$.

\bigskip

\noindent
\textsc{K29$^*$ algorithm}

\noindent
\textbf{Parameter:} forecast-continuous kernel $\KKK$ on $[0,1]\times\mathbf{X}$

\parshape=8
\IndentI   \WidthI
\IndentII  \WidthII
\IndentII  \WidthII
\IndentIII \WidthIII
\IndentII  \WidthII
\IndentIII \WidthIII
\IndentII  \WidthII
\IndentI   \WidthI
\noindent
FOR $n=1,2,\ldots$:\\
  Read $x_n\in\mathbf{X}$.\\
  Set $S_n(p) := \sum_{i=1}^{n-1} \KKK((p,x_n),(p_i,x_i))(y_i-p_i)
    +\frac12\KKK((p,x_n),(p,x_n))(1-2p)$\\
    for $p\in[0,1]$.\\
  If $\sign S_n(0)=\sign S_n(1)\ne0$, output $p_n:=(1+\sign S_n(0))/2$;\\
    otherwise, output any root $p$ of $S_n(p)=0$ as $p_n$.\\
  Read $y_n\in\{0,1\}$.\\
END FOR.

\bigskip

\noindent
(Since the function $S_n(p)$ is continuous,
the equation $S_n(p)=0$ indeed has a solution
when $\sign S_n(0)=\sign S_n(1)\ne0$ does not hold;
remember that $\sign S$ is $1$ for $S$ positive,
$-1$ for $S$ negative, and $0$ for $S=0$.)
The main term in the expression for $S_n(p)$
is $\sum_{i=1}^{n-1} \KKK((p,x_n),(p_i,x_i))(y_i-p_i)$.
Ignoring the other term for a moment,
we can describe the intuition behind this algorithm
by saying that $p_n$ is chosen so that $p_i$ are unbiased forecasts for $y_i$
on the rounds $i=1,\ldots,n-1$ for which $(p_i,x_i)$ is similar to $(p_n,x_n)$.
The term $\frac12\KKK((p,x_n),(p,x_n))(1-2p)$,
which can be rewritten as
$
  \KKK((p,x_n),(p,x_n))(0.5-p)
$,
adds an element of regularization,
i.e., bias towards the ``neutral'' value $p_n=0.5$.

The K29$^*$ algorithm requires solving the equation $S_n(p)=0$,
but this can be easily done using the bisection method
or one of the numerous more sophisticated methods
(see, e.g., \cite{press/etal:1992}, Chapter 9).

It is well known
(see \cite{doob:1953}, Theorem~II.3.1, for a simple proof)
that there exists a function $\Phi:[0,1]\times\mathbf{X}\to\HHH$
(a \emph{feature mapping} taking values in a Hilbert space\footnote
  {Hilbert spaces in this paper are allowed to be non-separable or finite dimensional;
  we, however, always assume that their dimension is at least $1$.}
$\HHH$ called the \emph{feature space})
such that
\begin{equation}\label{eq:K1}
  \KKK(a,b)
  =
  \left\langle
    \Phi(a),\Phi(b)
  \right\rangle_{\HHH},
  \enspace
  \forall a,b\in[0,1]\times\mathbf{X}
\end{equation}
($\langle\cdot,\cdot\rangle_{\HHH}$ standing for the inner product in $\HHH$).
It is known that,
for any $\KKK$ and $\Phi$ connected by (\ref{eq:K1}),
$\KKK$ is forecast-continuous if and only if
$\Phi$ is a continuous function of $p$ for each fixed $x\in\mathbf{X}$
(see Appendix \ref{app:continuity}).

Now we can state the basic result about K29$^*$
(proved in Appendix \ref{app:K29}).
\begin{theorem}\label{thm:K29}
  Let $\KKK$ be the kernel defined by~(\ref{eq:K1})
  for a  feature mapping $\Phi:[0,1]\times\mathbf{X}\to\HHH$
  continuous in its first argument.
  The K29\/$^*$ algorithm with parameter $\KKK$ ensures
  \begin{multline}\label{eq:K29}
    \left\|
      \sum_{n=1}^N
      (y_n-p_n)
      \Phi(p_n,x_n)
    \right\|^2_{\HHH}
    \le
    \sum_{n=1}^N
    p_n(1-p_n)
    \left\|
      \Phi(p_n,x_n)
    \right\|^2_{\HHH},
    \\
    \forall N\in\{1,2,\ldots\}.
  \end{multline}
\end{theorem}

Let us assume, for simplicity, that
\begin{equation}\label{eq:C}
  \CC_{\KKK}
  :=
  \sup_{p,x}
  \left\|
    \Phi(p,x)
  \right\|_{\HHH}
  <
  \infty
\end{equation}
(it is often a good idea to use kernels with $\left\|\Phi(p,x)\right\|_{\HHH}\equiv1$
and, therefore, $\CC_{\KKK}=1$).
Equation~(\ref{eq:K29}) then implies
\begin{equation}\label{eq:simple}
  \left\|
    \sum_{n=1}^N
    (y_n-p_n)
    \Phi(p_n,x_n)
  \right\|_{\HHH}
  \le
  \frac{\CC_{\KKK}}{2}
  \sqrt{N},
  \quad
  \forall N\in\{1,2,\ldots\}.
\end{equation}
When $\Phi$ is absent (in the sense $\Phi\equiv1$),
this shows that the forecasts $p_n$ are unbiased,
in the sense that they are close to $y_n$ on average;
the presence of $\Phi$ implies, for a suitable kernel,
``local unbiasedness''.
This is further discussed in the first part of \S\ref{sec:discussion}.

In the conference version \cite{\GTPXIII} of this paper
we also considered the K29 algorithm,
which differs from K29$^*$ in that $S_n(p)$ is defined as
\begin{equation*}
  S_n(p)
  :=
  \sum_{i=1}^{n-1} \KKK((p,x_n),(p_i,x_i))(y_i-p_i)
\end{equation*}
and that the requirement that $\KKK$ should be forecast-continuous is slightly relaxed
(the joint continuity in $(p,p')$ is replaced by the separate continuity in $p$ and $p'$).
For the K29 algorithm,
the inequality (\ref{eq:K29}) continues to hold
if $p_n(1-p_n)$ is removed;
therefore, (\ref{eq:simple}) continues to hold
if the denominator $2$ is removed.
We will sometimes use ``algorithms of large numbers''
as generic name for the K29 and K29$^*$ algorithms;
the motivation for these names is that the main properties of these algorithms
are easy corollaries of Kolmogorov's 1929 proof \cite{kolmogorov:1929LLN}
of the weak law of large numbers.

\section{Reproducing kernel Hilbert spaces}
\label{sec:FS}

A \emph{reproducing kernel Hilbert space} (RKHS) on a set $Z$
is a Hilbert space $\FFF$ of real-valued functions on $Z$
such that the evaluation functional $f\in\FFF\mapsto f(z)$
is continuous for each $z\in Z$.
By the Riesz--Fischer theorem,
for each $z\in Z$ there exists a function $\KKK_z\in\FFF$ such that
\begin{equation}\label{eq:RF}
  f(z)
  =
  \langle \KKK_z,f\rangle_{\FFF},
  \quad
  \forall f\in\FFF.
\end{equation}
The \emph{kernel of RKHS $\FFF$} is
\begin{equation}\label{eq:K2}
  \KKK(z,z')
  :=
  \left\langle
    \KKK_z,\KKK_{z'}
  \right\rangle_{\FFF}
\end{equation}
(equivalently, we could define $\KKK(z,z')$ as $\KKK_z(z')$
or as $\KKK_{z'}(z)$).
Since (\ref{eq:K2}) is a special case of (\ref{eq:K1}),
the function $\KKK$ defined by (\ref{eq:K2}) is indeed a kernel on $Z$,
as defined earlier.
On the other hand,
for every kernel $\KKK$ on $Z$
there exists a unique RKHS $\FFF$ on $Z$
such that $\KKK$ is the kernel of $\FFF$
(see, e.g., \cite{aronszajn:1944}, Theorem 2).

A long list of RKHS and the corresponding kernels is given
in \cite{berlinet/thomas-agnan:2004}, \S7.4.
Perhaps the most interesting RKHS in our current context
are various Sobolev spaces $W^{m,p}(\Omega)$
(\cite{adams/fournier:2003} is the standard reference for the latter).
We will be interested in the especially simple space $W^{1,2}([0,1])$,
to be defined shortly;
but first let us make a brief terminological remark.
The term ``Sobolev space'' is usually treated as the name for a topological vector space.
All these spaces are normable,
but different norms are not considered to lead
to different Sobolev spaces
as long as the topology does not change.

The \emph{Fermi--Sobolev norm}
$\left\|f\right\|_{\FS}$ of a smooth function $f:[0,1]\to\bbbr$
is defined by
\begin{equation}\label{eq:norm}
  \left\|f\right\|_{\FS}^2
  :=
  \left(
    \int_0^1
      f(t)
    \D t
  \right)^2
  +
  \int_0^1
    \left(
      f'(t)
    \right)^2
  \D t.
\end{equation}
The \emph{Fermi--Sobolev space} on $[0,1]$ is the completion
of the set of smooth $f:[0,1]\to\bbbr$
satisfying
$\left\|f\right\|_{\FS} < \infty$
with respect to the norm $\left\|\cdot\right\|_{\FS}$.
It is easy to see that it is in fact an RKHS
(indeed, if $\left\|f\right\|_{\FS}=c<\infty$,
the mean of $f$ is bounded by $c$ in absolute value
and $\left|f(b)-f(a)\right|\le\int_a^b\left|f'(t)\right|\D t\le c$
for all $0\le a<b\le1$).
As a topological vector space,
it coincides with the Sobolev space $W^{1,2}([0,1])$.
The \emph{Fermi--Sobolev space} on $[0,1]^k$
is the tensor product of $k$ copies of the Fermi--Sobolev space on $[0,1]$.

The kernel of the Fermi--Sobolev space on $[0,1]$
was found in \cite{craven/wahba:1979}
(see also \cite{wahba:1990}, \S10.2);
it is given by
\begin{align}
  \KKK(t,t')
  &=
  k_0(t) k_0(t')
  +
  k_1(t) k_1(t')
  +
  k_2(\lvert t-t'\rvert)\notag\\
  &=
  1
  +
  \left(
    t-\frac12
  \right)
  \left(
    t'-\frac12
  \right)
  +
  \frac12
  \left(
    \lvert t-t' \rvert^2
    -
    \lvert t-t' \rvert
    +
    \frac16
  \right)\notag\\
  &=
  \frac12
  \minsq(t,t')
  +
  \frac12
  \minsq(1-t,1-t')
  +
  \frac56,
  \label{eq:final}
\end{align}
where $k_l:=B_l/l!$ are scaled Bernoulli polynomials $B_l$.
\ifarXiv
  We will derive the final expression for $\KKK(t,t')$ in (\ref{eq:final})
  in Appendix~\ref{app:kernel}.
\fi
For the Fermi--Sobolev space on $[0,1]^k$
we have
\begin{equation}\label{eq:FSkernel}
  \KKK
  \left(
    (t_1,\ldots,t_k),
    (t'_1,\ldots,t'_k)
  \right)
  =
  \prod_{i=1}^k
  \left(
    \frac12
    \minsq(t_i,t'_i)
    +
    \frac12
    \minsq(1-t_i,1-t'_i)
    +
    \frac56
  \right)
\end{equation}
and, therefore,
\begin{equation}\label{eq:c-k}
  \CC^2_{\KKK}
  =
  \max_{t\in[0,1]}
  \left(
    \frac12
    t^2
    +
    \frac12
    (1-t)^2
    +
    \frac56
  \right)^k
  =
  \left(
    \frac43
  \right)^k.
\end{equation}
For further information about the Fermi--Sobolev spaces,
see
\ifJOURNAL
  \cite{\GTPXIII}
  (especially the technical report).
\fi
\ifarXiv
  \cite{\GTPXIII}.
\fi
\ifWP
  \cite{\GTPXIII}.
\fi

\section{The K29$^*$ algorithm in RKHS}
\label{sec:N91}

We can now deduce the following corollary
from Theorem~\ref{thm:K29}.
\begin{theorem}\label{thm:N91}
  Let $\FFF$ be an RKHS on $[0,1]\times\mathbf{X}$
  with a forecast-continuous kernel $\KKK$.
  The K29\/$^*$ algorithm with parameter $\KKK$ ensures
  \begin{equation}\label{eq:N91}
    \left|
      \sum_{n=1}^N
      (y_n-p_n)
      f(p_n,x_n)
    \right|
    \le
    \left\|f\right\|_{\FFF}
    \sqrt{\sum_{n=1}^N p_n(1-p_n) \KKK((p_n,x_n),(p_n,x_n))}
  \end{equation}
  for all $N$ and all $f\in\FFF$.
\end{theorem}

\ifFULL
To prove this theorem
we will need some further properties of RKHS.
[It seems in fact this is not needed.]
Earlier we discussed
two ways to introduce the notions
of an RKHS and a kernel:
one can start from the former
or from the latter.
A popular third way is to start from a picture
that involves both notions,
closely intertwined.
A function $\KKK:Z^2\to\bbbr$
is said to be a \emph{reproducing kernel}
of a Hilbert space $\FFF$ of functions on $Z$
if:
\begin{itemize}
\item
  for every $z\in Z$,
  \begin{equation}\label{eq:reproducing1}
    \KKK(\cdot,z)
    \in
    \FFF;
  \end{equation}
\item
  for all $f\in\FFF$ and $z\in Z$,
  \begin{equation}\label{eq:reproducing2}
    f(z)
    =
    \langle
      f(\cdot),\KKK(\cdot,z)
    \rangle_{\FFF}.
  \end{equation}
\end{itemize}
Since a Hilbert function space
can never have more than one reproducing kernel
(\cite{aronszajn:1950}, \S I.2, (1)),
we will say that such a $\KKK$
is \emph{the} reproducing kernel of $\FFF$.

All three ways are equivalent
in the following sense:
\begin{itemize}
\item
  if $\FFF$ is an RKHS on $Z$,
  its kernel $\KKK$ is a kernel on $Z$
  satisfying (\ref{eq:reproducing1}) and (\ref{eq:reproducing2})
  (see (\ref{eq:RF}));
\item
  if $\FFF$ is a Hilbert space $\FFF$ of functions on $Z$
  with reproducing kernel $\KKK$,
  then $\KKK$ is a kernel on $Z$
  (\cite{aronszajn:1944}, \S I.2, (2,2) and (2,4))
  and $\FFF$ is an RKHS
  (\cite{aronszajn:1944}, Th\'eor\`eme 1)
  with kernel $\KKK$
  (this follows immediately from (\ref{eq:reproducing2}));
\item
  if $\KKK$ is a kernel on $Z$,
  there is one and only one Hilbert function space $\FFF$
  with $\KKK$ as its reproducing kernel
  (\cite{aronszajn:1944}, Th\'eor\`eme 2)
  and
  there is one and only one RKHS $\FFF$
  with $\KKK$ as its kernel
  (this follows from the previous statements).
\end{itemize}
\fi

\begin{proof}
  Applying K29$^*$ to the feature mapping
  $
    (p,x)\in[0,1]\times\mathbf{X}
    \mapsto
    \KKK_{p,x}\in\FFF
  $
  and using (\ref{eq:K29}),
  we obtain,
  for any $f\in\FFF$:
  \begin{multline*}
    \left|
      \sum_{n=1}^N
      (y_n-p_n)
      f(p_n,x_n)
    \right|
    =
    \left|
      \sum_{n=1}^N
      (y_n-p_n)
      \left\langle
        \KKK_{p_n,x_n}, f
      \right\rangle_{\FFF}
    \right|\\
    =
    \left|
      \left\langle
        \sum_{n=1}^N
        (y_n-p_n)
        \KKK_{p_n,x_n},
        f
      \right\rangle_{\FFF}
    \right|
    \le
    \left\|
      \sum_{n=1}^N
      (y_n-p_n)
      \KKK_{p_n,x_n}
    \right\|_{\FFF}
    \left\|
      f
    \right\|_{\FFF}\\
    \le
    \left\|
      f
    \right\|_{\FFF}
    \sqrt
    {
      \sum_{n=1}^N
      p_n(1-p_n)
      \KKK((p_n,x_n),(p_n,x_n))
    }.
    \qedmath
  \end{multline*}
\end{proof}

When $\CC_{\KKK}$ in (\ref{eq:C}) is finite,
(\ref{eq:N91}) implies
\begin{equation}\label{eq:N91simple}
  \left|
    \sum_{n=1}^N
    (y_n-p_n)
    f(p_n,x_n)
  \right|
  \le
  \frac{\CC_{\KKK}}{2}
  \left\|f\right\|_{\FFF}
  \sqrt{N}.
\end{equation}

\section{Informal discussion}
\label{sec:discussion}

In this section we explain
why the inequalities in Theorems \ref{thm:K29} and \ref{thm:N91}
can be interpreted as results about calibration and resolution,
and then briefly discuss a puzzling aspect of the algorithms of large numbers.
For concreteness,
we usually talk about the K29$^*$ algorithm,
but all we say can also be applied, with obvious modifications,
to K29.

\subsection*{Calibration, resolution, and calibration-cum-resolution}


We start from the intuitive notion of calibration
(for further details, see \cite{dawid:1986} and \cite{foster/vohra:1998}).
The forecasts $p_n$, $n=1,\ldots,N$, are said to be ``well calibrated''
(or ``unbiased in the small'', or ``reliable'', or ``valid'')
if, for any $p^*\in[0,1]$,
\begin{equation}\label{eq:p1}
  \frac
  {\sum_{n=1,\ldots,N:p_n\approx p^*} y_n}
  {\sum_{n=1,\ldots,N:p_n\approx p^*} 1}
  \approx
  p^*
\end{equation}
provided $\sum_{n=1,\ldots,N:p_n\approx p^*} 1$ is not too small.
The interpretation of (\ref{eq:p1}) is that the forecasts
should be in agreement with the observed frequencies.
It will be convenient to rewrite (\ref{eq:p1}) as
\begin{equation}\label{eq:p2}
  \frac
  {\sum_{n=1,\ldots,N:p_n\approx p^*} (y_n-p_n)}
  {\sum_{n=1,\ldots,N:p_n\approx p^*} 1}
  \approx
  0.
\end{equation}

The fact that good calibration is only a necessary condition
for good forecasting performance
can be seen from the following standard example
\cite{dawid:1986,foster/vohra:1998}:
if
\begin{equation*}
  (y_1,y_2,y_3,y_4,\ldots)
  =
  (1,0,1,0,\ldots),
\end{equation*}
the forecasts $p_n=1/2$, $n=1,2,\ldots$,
are well calibrated but rather poor;
it would be better to forecast with
\begin{equation*}
  (p_1,p_2,p_3,p_4,\ldots)
  =
  (1,0,1,0,\ldots).
\end{equation*}
Assuming that each datum $x_n$ contains the information about the parity of $n$
(which can always be added to $x_n$),
we can see that the problem with the forecasting strategy $p_n\equiv1/2$
is its lack of resolution:
it does not distinguish between the data with odd and even $n$.
In general, we would like each forecast $p_n$
to be as specific as possible to the current datum $x_n$;
the resolution of a forecasting algorithm
is the degree to which it achieves this goal
(taking it for granted that $x_n$ contains all relevant information).

Analogously to (\ref{eq:p2}),
the forecasts $p_n$, $n=1,\ldots,N$, may be said to have good resolution
if, for any $x^*\in\mathbf{X}$,
\begin{equation*}
  \frac
  {\sum_{n=1,\ldots,N:x_n\approx x^*} (y_n-p_n)}
  {\sum_{n=1,\ldots,N:x_n\approx x^*} 1}
  \approx
  0
\end{equation*}
provided the denominator is not too small.
We can also require that the forecasts $p_n$, $n=1,\ldots,N$,
should have good ``calibration-cum-resolution'':
for any $(p^*,x^*)\in[0,1]\times\mathbf{X}$,
\begin{equation*}
  \frac
  {\sum_{n=1,\ldots,N:(p_n,x_n)\approx(p^*,x^*)} (y_n-p_n)}
  {\sum_{n=1,\ldots,N:(p_n,x_n)\approx(p^*,x^*)} 1}
  \approx
  0
\end{equation*}
provided the denominator is not too small.
Notice that even if forecasts have both good calibration and good resolution,
they can still have poor calibration-cum-resolution.

It is easy to see that (\ref{eq:simple}) implies good calibration-cum-resolution
for a suitable $\Phi$ and large $N$:
indeed, (\ref{eq:simple}) shows that the forecasts $p_n$ are unbiased
in the neighborhood of each $(p^*,x^*)$
for functions $\Phi$ that map distant $(p,x)$ and $(p',x')$
to almost orthogonal elements of the feature space
(such as $\Phi$ corresponding to the Gaussian kernel
\begin{equation}\label{eq:width}
  \KKK
  \left(
    (p,x),(p',x')
  \right)
  :=
  \exp
  \left(
    \frac{(p-p')^2+\left\|x-x'\right\|^2}{2\sigma^2}
  \right)
\end{equation}
for a small ``kernel width'' $\sigma>0$).

In general,
to make sense of the $\approx$
in the numerator and denominator
of, say, (\ref{eq:p2}),
we replace each ``crisp'' point $p^*$
by a ``fuzzy point'' $I_{p^*}:[0,1]\to[0,1]$;
$I_{p^*}$ is required to be continuous,
and we might also want to have $I_{p^*}(p^*)=1$
and $I_{p^*}(p)=0$ for all $p$ outside a small neighborhood of $p^*$.
The alternative of choosing $I_{p^*}:=\III_{[p_{-},p_{+}]}$,
where $[p_{-},p_{+}]$ is a short interval containing $p^*$
and $\III_{[p_{-},p_{+}]}$ is its indicator function,
does not work because of Oakes's and Dawid's examples
\cite{oakes:1985,dawid:1985JASA};
$I_{p^*}$ can, however,
be arbitrarily close to $\III_{[p_{-},p_{+}]}$.

Consider, e.g., the following approximation to the indicator function
of a short interval $[p_{-},p_{+}]$
containing $p^*$:
\begin{equation}\label{eq:neighborhood}
  f(p)
  :=
  \begin{cases}
    1 & \text{if $p_{-}+\epsilon \le p \le p_{+}-\epsilon$}\\
    0 & \text{if $p \le p_{-}-\epsilon$ or $p \ge p_{+}+\epsilon$}\\
    \frac12 + \frac{1}{2\epsilon}(p-p_{-}) & \text{if $p_{-}-\epsilon \le p \le p_{-}+\epsilon$}\\
    \frac12 + \frac{1}{2\epsilon}(p_{+}-p) & \text{if $p_{+}-\epsilon \le p \le p_{+}+\epsilon$};
  \end{cases}
\end{equation}
we assume that $\epsilon>0$ satisfies
\begin{equation*}
  0 < p_{-}-\epsilon < p_{-}+\epsilon < p_{+}-\epsilon < p_{+}+\epsilon < 1.
\end{equation*}
It is clear that this approximation belongs to the Fermi--Sobolev space.
An easy computation shows that (\ref{eq:N91simple}) and (\ref{eq:c-k}) imply
\begin{equation}\label{eq:N91calibration}
  \left|
    \sum_{n=1}^N
    (y_n-p_n)
    f(p_n)
  \right|
  \le
  \frac{1}{\sqrt{3}}
  \sqrt
  {
    \left(
      \frac{1}{\epsilon}
      +
      (p_{+}-p_{-})^2
    \right)
    N
  }
\end{equation}
for all $N$.
We can see that (\ref{eq:p2}),
in the form
\begin{equation*}
  \frac
  {\sum_{n=1,\ldots,N} f(p_n)(y_n-p_n)}
  {\sum_{n=1,\ldots,N} f(p_n)}
  \approx
  0,
\end{equation*}
will hold if
\begin{equation*}
  \sum_{n=1}^N
  f(p_n)
  \gg
  \sqrt{N}
\end{equation*}
(roughly,
if significantly more than $\sqrt{N}$ forecasts
fall in the neighborhood $[p^{-},p^{+}]$ of $p^*$).

It is clear that inequalities analogous to (\ref{eq:N91calibration})
can also be proved for ``soft neighborhoods'' of points $(p^*,x^*)$
in $[0,1]\times\mathbf{X}$
(at least when $\mathbf{X}$ is a domain in a Euclidean space),
and so Theorem~\ref{thm:N91} also implies good calibration-cum-resolution for large $N$.
Convenient neighborhoods in $[0,1]\times[0,1]^{K}$ can be constructed
as tensor products of neighborhoods (\ref{eq:neighborhood}).

Inequality (\ref{eq:N91calibration})
and analogous inequalities expressing resolution and calibration-cum-resolution
are explicit in the sense that they do not involve limits, $o$, $O$,
unspecified constants, etc.
The price to pay is their relative complexity;
therefore, we also state a simple asymptotic result about calibration-cum-resolution.
\begin{corollary}\label{cor:asymptotic}
  If $\mathbf{X}$ is a compact metric space,
  some forecasting algorithm guarantees
  \begin{equation}\label{eq:asymptotic}
    \lim_{N\to\infty}
    \frac1N
    \sum_{n=1}^N
    (y_n-p_n)
    f(p_n,x_n)
    =
    0
  \end{equation}
  for all continuous functions $f:[0,1]\times\mathbf{X}\to\bbbr$.
\end{corollary}
Calibration corresponds
to the case where $f(p,x)=I_{p^*}(p)$ does not depend on $x$
and resolution to the case where $f(p,x)=I_{x^*}(x)$ does not depend on $p$.
This result was proved in \cite{kakade/foster:2004}
in the case of calibration (there are no $x_n$) and Lipschitz functions $f$.
\begin{Proof}{of Corollary \ref{cor:asymptotic}}
  Let $\FFF$ be an RKHS on $[0,1]\times\mathbf{X}$
  which is \emph{universal}, i.e., dense in the space $C([0,1]\times\mathbf{X})$,
  and whose kernel $\KKK$ is continuous and satisfies $\CC_{\KKK}<\infty$.
  The notion of universality is introduced in \cite{steinwart:2001}, Definition 4,
  and the existence of such an $\FFF$ is shown in \cite{steinwart/etal:2006}, Theorem 2.
  For any continuous function $f:[0,1]\times\mathbf{X}\to\bbbr$
  there is a $g\in\FFF$ that is $\epsilon$-close to $f$ in the metric $C([0,1]\times\mathbf{X})$,
  and so, by (\ref{eq:N91simple}),
  \begin{multline*}
    \limsup_{N\to\infty}
    \left|
      \frac1N
      \sum_{n=1}^N
      (y_n-p_n)
      f(p_n,x_n)
    \right|
    \le
    \limsup_{N\to\infty}
    \left|
      \frac1N
      \sum_{n=1}^N
      (y_n-p_n)
      g(p_n,x_n)
    \right|
    +
    \epsilon\\
    \le
    \limsup_{N\to\infty}
    \frac1N
    \frac{\CC_{\KKK}}{2}
    \left\|g\right\|_{\FFF}
    \sqrt{N}
    +
    \epsilon
    =
    \epsilon;
  \end{multline*}
  since this holds for any $\epsilon>0$,
  (\ref{eq:asymptotic}) also holds.
  \qedtext
\end{Proof}
One of the algorithms achieving (\ref{eq:asymptotic})
for $\mathbf{X}=[0,1]^k$
is K29$^*$ applied to the Fermi--Sobolev kernel (\ref{eq:FSkernel}).
It is interesting, and somewhat counterintuitive,
that K29$^*$ applied to the Gaussian kernel (\ref{eq:width})
(with any $\sigma>0$)
also achieves (\ref{eq:asymptotic});
the universality of the Gaussian kernels is proved in \cite{steinwart:2001}
(Example 1).

\ifFULL
Let $(p^*,x^*)$ be a point in $[0,1]\times\mathbf{X}$.
Fix a forecast-continuous kernel $\KKK:([0,1]\times\mathbf{X})^2\to\bbbr$
and consider the ``soft neighborhood''
\begin{equation}\label{eq:I}
  I_{(p^*,x^*)}(p,x)
  :=
  \KKK((p^*,x^*),(p,x))
\end{equation}
of the point $(p^*,x^*)$.
The following is an easy corollary of Theorem~\ref{thm:K29}
(and a special case of Theorem \ref{thm:N91}).
\begin{corollary}\label{cor:calibration}
  The K29$^*$ algorithm with parameter $\KKK$ such that
  \begin{equation*}
    \CC_{\KKK}^2
    :=
    \sup_{(p,x)\in[0,1]\times\mathbf{X}}
    \KKK((p,x),(p,x))
    <
    \infty
  \end{equation*}
  ensures
  \begin{equation}\label{eq:calibration}
    \left|
      \sum_{n=1}^N
      (y_n-p_n)
      I_{(p^*,x^*)}(p_n,x_n)
    \right|
    \le
    \frac{\CC_{\KKK}^2}{2}
    \sqrt{N}
  \end{equation}
  for each point $(p^*,x^*)\in[0,1]\times\mathbf{X}$,
  where $I$ is defined by (\ref{eq:I}).
\end{corollary}
\begin{proof}
  Let $\Phi:[0,1]\times\mathbf{X}\to\HHH$ be a function
  taking values in a Hilbert space $\HHH$
  and satisfying (\ref{eq:K1});
  the constant $\CC_{\KKK}$ can be equivalently defined by (\ref{eq:C}).
  The Cauchy--Schwarz inequality, (\ref{eq:C}), and~(\ref{eq:simple}) imply
  \begin{multline*}
    \left|
      \sum_{n=1}^N
      (y_n-p_n)
      I_{(p^*,x^*)}(p_n,x_n)
    \right|\\
    =
    \left|
      \left(
        \sum_{n=1}^N
        (y_n-p_n)
        \Phi(p_n,x_n)
      \right)
      \cdot
      \Phi(p^*,x^*)
    \right|\\
    \le
    \left\|
      \sum_{n=1}^N
      (y_n-p_n)
      \Phi(p_n,x_n)
    \right\|_{\HHH}
    \left\|
      \Phi(p^*,x^*)
    \right\|_{\HHH}
    \le
    \frac{\CC_{\KKK}^2}{2}
    \sqrt{N}.
    \qedmath
  \end{multline*}
\end{proof}

Taking as $\KKK$ a non-negative forecast-continuous kernel that ignores the data,
\begin{equation*}
  \KKK((p,x),(p',x'))
  =
  \KKK(p,p'),
\end{equation*}
we can rewrite~(\ref{eq:calibration}) as
\begin{equation*}
  \left|
    \frac
    {
      \sum_{n=1}^N
      (y_n-p_n)
      I_{p^*}(p_n)
    }
    {
      \sum_{n=1}^N
      I_{p^*}(p_n)
    }
  \right|
  \le
  \frac
  {
    \CC_{\KKK}^2\sqrt{N}
  }
  {
    \sum_{n=1}^N
    I_{p^*}(p_n)
  };
\end{equation*}
comparing this with~(\ref{eq:p2}),
we can see that K29$^*$ is well calibrated provided
\begin{equation*}
  \sum_{n=1}^N
  I_{p^*}(p_n)
  \gg
  \sqrt{N}.
\end{equation*}
\fi

Our discussion of calibration and resolution
in this subsection
has been somewhat speculative,
and the reader might ask whether these two properties are really useful.
This question is answered, to some degree, in \cite{\GTPXIV,\GTPXVII},
which show that probability forecasts satisfying these properties
lead to good decisions
(at least in the simple decision protocols considered in those papers).

\subsection*{Puzzle of the iterated logarithm}

Theorems~\ref{thm:K29} and \ref{thm:N91} imply
that the forecasts produced by the K29$^*$ algorithm
are even closer to the actual observations on average
than in the case of ``genuine randomness'',
where Reality produces the data and observations
from a probability distribution on $(\mathbf{X}\times\{0,1\})^{\infty}$
and each $p_n$ is the conditional probability that $y_n=1$
given $x_1,\ldots,x_n$, $y_1,\ldots,y_{n-1}$,
and whatever further information may be available at this point.
Indeed, let us take, for simplicity, $\Phi\equiv1$ (and $\HHH:=\bbbr$)
in Theorem~\ref{thm:K29}.
According to the martingale law of the iterated logarithm
(see, e.g., \cite{stout:1970} or Chapter 5 of \cite{shafer/vovk:2001}),
we would expect
\begin{equation}\label{eq:LIL}
  \limsup_{N\to\infty}
  \frac
  {
    \left|
      \sum_{n=1}^N
      (y_n-p_n)
    \right|
  }
  {
    \sqrt{2A_N\ln\ln A_N}
  }
  =
  1,
\end{equation}
where
$
  A_N
  :=
  \sum_{n=1}^N
  p_n(1-p_n)
$
is assumed to tend to $\infty$ as $N\to\infty$,
and so expect, contrary to (\ref{eq:simple}),
\begin{equation*}
  \sup_{N\in\{1,2,\ldots\}}
  \frac
  {
    \left\|
      \sum_{n=1}^N
      (y_n-p_n)
      \Phi(p_n,x_n)
    \right\|_{\HHH}
  }
  {
    \sqrt{N}
  }
\end{equation*}
to be infinite for $p_n$ not consistently very close to 0 or 1.
Actually, in this case ($\Phi\equiv1$)
Forecaster can even make sure that
\begin{equation*}
  \left\|
    \sum_{n=1}^N
    (y_n-p_n)
    \Phi(p_n,x_n)
  \right\|_{\HHH}
  =
  \frac12,
  \quad
  \forall N\in\{1,2,\ldots\}
\end{equation*}
(choosing $p_1:=1/2$ and $p_n:=y_{n-1}$, $n=2,3,\ldots$).

For a general $\Phi$,
we can also expect that the probabilities $p_n$
contrived by the algorithms of large numbers
(K29 or K29$^*$)
will have better calibration and resolution than the true probabilities.
There is, however, little doubt that the true probabilities
are more useful than any probabilities we are able to come up with.
The true probabilities are not as good at calibration and resolution,
so they must be better in some other equally important respects.
It remains unclear what these other respects may be,
and this is what we call the puzzle of the iterated logarithm.

\section{Optimality of the K29$^*$ algorithm}
\label{sec:optimality}

In this section we establish that the inequalities
in Theorems \ref{thm:K29} and \ref{thm:N91}
are tight, in a natural sense.

\begin{figure}[bt]
  \centering
  \makebox{\includegraphics[width=5cm,angle=270]{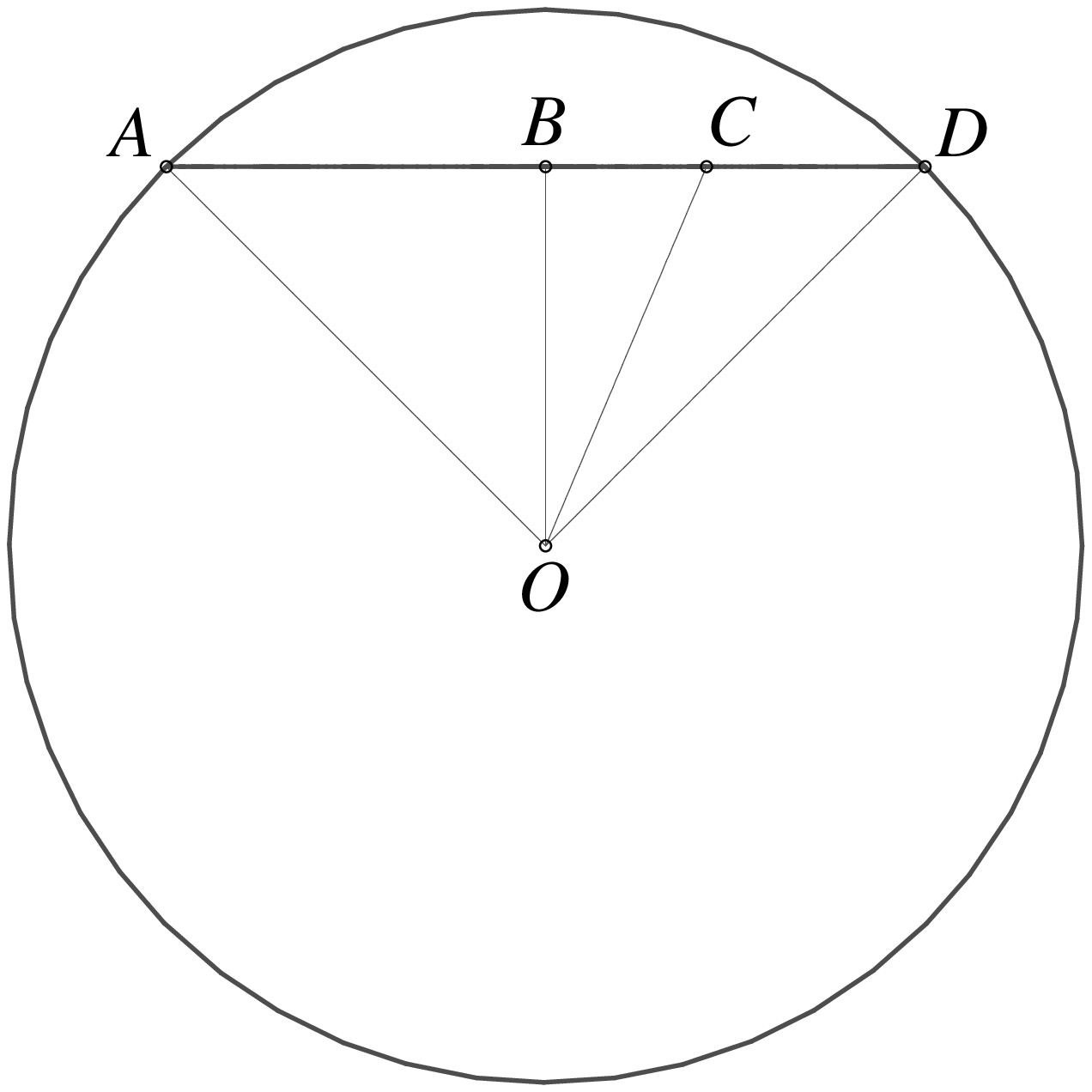}}
  \caption{\label{fig:worst}The worst case for Reality II;
    $\lvert OA\rvert=\lvert OD\rvert=R_N$,
    $\lvert OC\rvert=R_{N-1}$,
    $\lvert AC\rvert=(1-p_N)\Phi_N$,
    $\lvert CD\rvert=p_N\Phi_N$,
    $\lvert OB\rvert=h$.}
\end{figure}

Equation (\ref{eq:K29})
says that the differences $y_n-p_n$ are small on average,
even when scattered in a Hilbert space
by multiplying by $\Phi(p_n,x_n)$.
The next result says that it is the best Forecaster can do.
\begin{theorem}\label{thm:antiK29}
  Let $\Phi:[0,1]\times\mathbf{X}\to\HHH$,
  where $\HHH$ is a Hilbert space.
  There is a strategy for Reality II
  which guarantees that
  \begin{equation}\label{eq:antiK29}
    \left\|
      \sum_{n=1}^N
      (y_n-p_n)
      \Phi(p_n,x_n)
    \right\|^2_{\HHH}
    \ge
    \sum_{n=1}^N
    p_n(1-p_n)
    \left\|
      \Phi(p_n,x_n)
    \right\|^2_{\HHH}
  \end{equation}
  always holds for all $N=1,2,\ldots$,
  regardless of what the other players do.
\end{theorem}
\begin{proof}
  Set
  \begin{equation*}
    R_N
    :=
    \left\|
      \sum_{n=1}^N
      (y_n-p_n)
      \Phi(p_n,x_n)
    \right\|_{\HHH},
    \quad
    N=1,2,\ldots;
  \end{equation*}
  it is sufficient to show that on the $N$th round,
  $N=1,2,\ldots$,
  Reality II can ensure that
  \begin{equation}\label{eq:goal}
    R_N^2 - R_{N-1}^2
    \ge
    p_N(1-p_N)
    \Phi_N^2,
  \end{equation}
  where
  \begin{equation*}
    \Phi_N
    :=
    \left\|
      \Phi(p_N,x_N)
    \right\|_{\HHH}.
  \end{equation*}
  Fix an $N$.
  Define points $A,C,D\in\HHH$ as
  \begin{align*}
    C
    &:=
    \sum_{n=1}^{N-1}
    (y_n-p_n)
    \Phi(p_n,x_n),\\
    A
    &:=
    \sum_{n=1}^{N-1}
    (y_n-p_n)
    \Phi(p_n,x_n)
    +
    (1-p_N)
    \Phi(p_N,x_N),\\
    D
    &:=
    \sum_{n=1}^{N-1}
    (y_n-p_n)
    \Phi(p_n,x_n)
    +
    (-p_N)
    \Phi(p_N,x_N);
  \end{align*}
  it is up to Reality II whether make $R_N$ equal to $\lvert OA\rvert$ or $\lvert OD\rvert$,
  where $O$ is the origin.
  Assuming, without loss of generality, that $R_N=\max(\lvert OA\rvert,\lvert OD\rvert)$,
  we reduce our task to showing that the maximal value of $R_{N-1}$
  for fixed $R_N$, $\Phi_N$, and $p_N$ satisfies (\ref{eq:goal}).
  It is geometrically obvious
  (see the last paragraph of this proof for a rigorous argument)
  that $R_{N-1}$ attains its maximal value when
  $\lvert OA\rvert=\lvert OD\rvert$;
  this is illustrated in \Fig\ref{fig:worst}
  (remember that all four points, $O$, $A$, $C$, and $D$, lie in the same plane).
  Let $B$ be the base of the perpendicular dropped from $O$ onto the interval $AD$
  and $h:=\lvert OB\rvert$.
  Since the triangles $OBD$ and $OBC$ are right-angled,
  \begin{align*}
    R_N^2
    &=
    h^2
    +
    \left(
      \frac12 \Phi_N
    \right)^2,\\
    R_{N-1}^2
    &=
    h^2
    +
    \left(
      \frac12 \Phi_N
      -
      p_N \Phi_N
    \right)^2.
  \end{align*}
  Subtracting the second equality from the first,
  we obtain
  \begin{equation*}
    R_N^2
    -
    R_{N-1}^2
    =
    \left(
      \frac12 \Phi_N
    \right)^2
    -
    \left(
      \frac12 \Phi_N
      -
      p_N \Phi_N
    \right)^2
    =
    p_N(1-p_N)
    \Phi_N^2.
  \end{equation*}

  In conclusion,
  let us see that the maximum of $R_{N-1}$ is indeed attained when
  $\lvert OA\rvert=\lvert OD\rvert$.
  Assume that $\lvert OA\rvert=R_N$,
  with $\lvert OD\rvert$ now allowed to be less than $R_N$.
  Because of the compactness of the disk in \Fig\ref{fig:worst}
  (we are only interested in two-dimensional subspaces of $\HHH$,
  which are isometrically isomorphic to $\bbbr^2$),
  the maximum of $\lvert OC\rvert$ is attained at some point $C$.
  Supposing $\lvert OD\rvert<R_N$,
  it is, however, easy to check that no $C$
  will be a point of local maximum for $\lvert OC\rvert$;
  the least trivial case is perhaps
  where $O$ lies on the line $AD$ and $C$ is between $O$ and $D$.
  \qedtext
\end{proof}

The next result establishes the tightness of the bound in Theorem \ref{thm:N91}.
\begin{theorem}\label{thm:antiN91}
  Let $\FFF$ be an RKHS on $[0,1]\times\mathbf{X}$
  with kernel $\KKK$.
  Reality II has a strategy which ensures,
  regardless of what the other players do,
  that for each $N=1,2,\ldots$ there exists a non-zero $f\in\FFF$ such that
  \begin{equation}\label{eq:antiN91}
    \sum_{n=1}^N
    (y_n-p_n)
    f(p_n,x_n)
    \ge
    \left\|f\right\|_{\FFF}
    \sqrt{\sum_{n=1}^N p_n(1-p_n) \KKK((p_n,x_n),(p_n,x_n))}.
  \end{equation}
\end{theorem}
\begin{proof}
  By Theorem \ref{thm:antiK29} there exists a strategy for Reality II
  which ensures
  \begin{equation}\label{eq:easy}
    \left\|
      \sum_{n=1}^N
      (y_n-p_n)
      \KKK_{p_n,x_n}
    \right\|_{\FFF}
    \ge
    \sqrt
    {
      \sum_{n=1}^N
      p_n(1-p_n)
      \KKK((p_n,x_n),(p_n,x_n))
    }.
  \end{equation}
  Taking
  \begin{equation*}
    f
    :=
      \sum_{n=1}^N
      (y_n-p_n)
      \KKK_{p_n,x_n},
  \end{equation*}
  we obtain:
  \begin{multline*}
    \sum_{n=1}^N
    (y_n-p_n)
    f(p_n,x_n)
    =
    \sum_{n=1}^N
    (y_n-p_n)
    \left\langle
      \KKK_{p_n,x_n}, f
    \right\rangle_{\FFF}
    \\
    =
    \left\langle
      \sum_{n=1}^N
      (y_n-p_n)
      \KKK_{p_n,x_n},
      f
    \right\rangle_{\FFF}
    =
    \left\|
      \sum_{n=1}^N
      (y_n-p_n)
      \KKK_{p_n,x_n}
    \right\|_{\FFF}
    \left\|
      f
    \right\|_{\FFF}\\
    \ge
    \left\|
      f
    \right\|_{\FFF}
    \sqrt
    {
      \sum_{n=1}^N
      p_n(1-p_n)
      \KKK((p_n,x_n),(p_n,x_n))
    }.
  \end{multline*}
  If $f\ne0$,
  our task is accomplished.
  Otherwise, the right-hand side of (\ref{eq:easy}) will also be zero,
  and we can take any $f\ne0$.
  \qedtext
\end{proof}

\subsection*{Acknowledgments}

I am grateful to Ilia Nouretdinov for a discussion
that lead to the proof of Theorem \ref{thm:antiK29}
and to the anonymous reviewers of the conference and journal versions of this paper
for their comments.
This work was partially supported by MRC (grant S505/65)
and Royal Society.

\appendix
\section{Proof of Theorem~\ref{thm:K29}}
\label{app:K29}

The proof of Theorem~\ref{thm:K29}
is based on the game-theoretic approach
to the foundations of probability
proposed in~\cite{shafer/vovk:2001}.
A new player,
called Skeptic,
is added to the learning protocol of \S\ref{sec:K29};
the idea is that Skeptic is allowed to bet
at the odds defined by Forecaster's probabilities.
In this proof
there is no need to distinguish between Reality I and Reality II.

\bigskip

\noindent
\textsc{Binary Forecasting Game I}

\noindent
\textbf{Players:} Reality, Forecaster, Skeptic

\noindent
\textbf{Protocol:}

\parshape=8
\IndentI   \WidthI
\IndentI   \WidthI
\IndentII  \WidthII
\IndentII  \WidthII
\IndentII  \WidthII
\IndentII  \WidthII
\IndentII  \WidthII
\IndentI   \WidthI
\noindent
$\K_0 := C$.\\
FOR $n=1,2,\ldots$:\\
  Reality announces $x_n\in\mathbf{X}$.\\
  Forecaster announces $p_n\in[0,1]$.\\
  Skeptic announces $s_n\in\bbbr$.\\
  Reality announces $y_n\in\{0,1\}$.\\
  $\K_n := \K_{n-1} + s_n (y_n-p_n)$.\\
END FOR.

\bigskip

\noindent
The protocol describes not only the players' moves
but also the changes in Skeptic's capital $\K_n$;
its initial value is an arbitrary constant $C$.

The crucial (albeit very simple) observation
\cite{\GTPVIII}
is that for any continuous strategy for Skeptic
there exists a strategy for Forecaster
that does not allow Skeptic's capital to grow,
regardless of what Reality is doing
(similar observations were made in \cite{levin:1976uniformlatin} and \cite{kakade/foster:2004}).
To state this observation in its strongest form,
we will make Skeptic announce his strategy
for each round
before Forecaster's move on that round
rather than announce his full strategy at the beginning of the game.
Therefore,
we consider the following perfect-information game:

\bigskip

\noindent
\textsc{Binary Forecasting Game II}

\noindent
\textbf{Players:} Reality, Forecaster, Skeptic

\noindent
\textbf{Protocol:}

\parshape=8
\IndentI   \WidthI
\IndentI   \WidthI
\IndentII  \WidthII
\IndentII  \WidthII
\IndentII  \WidthII
\IndentII  \WidthII
\IndentII  \WidthII
\IndentI   \WidthI
\noindent
$\K_0 := C$.\\
FOR $n=1,2,\ldots$:\\
  Reality announces $x_n\in\mathbf{X}$.\\
  Skeptic announces continuous $S_n:[0,1]\to\bbbr$.\\
  Forecaster announces $p_n\in[0,1]$.\\
  Reality announces $y_n\in\{0,1\}$.\\
  $\K_n := \K_{n-1} + S_n(p_n) (y_n-p_n)$.\\
END FOR.

\bigskip

\begin{lemma}\label{lem:basic}
  Forecaster has a strategy in Binary Forecasting Game II
  that ensures $\K_0\ge\K_1\ge\K_2\ge\cdots$.
\end{lemma}
\begin{proof}
  Forecaster can use the following strategy to ensure $\K_0\ge\K_1\ge\cdots$:
  \begin{itemize}
  \item
    if $S_n(0)$ and $S_n(1)$ are both positive or both negative,
    take $p_n:=(1+\sign S_n(0))/2$;
  \item
    otherwise,
    choose $p_n$ so that $S_n(p_n)=0$
    (such a $p_n$ will exist).
    \qedtext
  \end{itemize}
\end{proof}

A measure-theoretic version of Lemma \ref{lem:basic}
(involving randomization)
was proved in \cite{sandroni:2003}, Proposition 1.

\subsection*{Proof of the theorem}

We start by noticing that
\begin{equation}\label{eq:expectation}
  (y_n-p_n)^2
  =
  p_n(1-p_n)
  +
  (1-2p_n)
  (y_n-p_n)
\end{equation}
both for $y_n=0$ and for $y_n=1$.
Following K29$^*$,
Forecaster ensures that Skeptic will never increase his capital
with the strategy
\begin{equation}\label{eq:strategy}
  s_n
  :=
  \sum_{i=1}^{n-1}
  \KKK
  \left(
    (p_n,x_n),(p_i,x_i)
  \right)
  (y_i-p_i)
  +
  \frac12
  \KKK
  \left(
    (p_n,x_n),(p_n,x_n)
  \right)
  (1-2p_n)
\end{equation}
(continuous in $p_n$ by our assumptions).
The increase in Skeptic's capital when he follows~(\ref{eq:strategy}) is
\begin{align*}
  \K_N-\K_0
  &=
  \sum_{n=1}^N
  s_n(y_n-p_n)\\
  &=
  \sum_{n=1}^N
  \sum_{i=1}^{n-1}
  \KKK
  \left(
    (p_n,x_n),(p_i,x_i)
  \right)
  (y_n-p_n)
  (y_i-p_i)\\
  &\quad{}+
  \frac12
  \sum_{n=1}^N
  \KKK
  \left(
    (p_n,x_n),(p_n,x_n)
  \right)
  (1-2p_n)
  (y_n-p_n)\\
  &=
  \frac12
  \sum_{n=1}^N
  \sum_{i=1}^N
  \KKK
  \left(
    (p_n,x_n),(p_i,x_i)
  \right)
  (y_n-p_n)
  (y_i-p_i)\\
  &\quad{}-
  \frac12
  \sum_{n=1}^N
  \KKK
  \left(
    (p_n,x_n),(p_n,x_n)
  \right)
  (y_n-p_n)^2\\
  &\quad{}+
  \frac12
  \sum_{n=1}^N
  \KKK
  \left(
    (p_n,x_n),(p_n,x_n)
  \right)
  (1-2p_n)
  (y_n-p_n)\\
  &=
  \frac12
  \sum_{n=1}^N
  \sum_{i=1}^N
  \KKK
  \left(
    (p_n,x_n),(p_i,x_i)
  \right)
  (y_n-p_n)
  (y_i-p_i)\\
  &\quad{}-
  \frac12
  \sum_{n=1}^N
  \KKK
  \left(
    (p_n,x_n),(p_n,x_n)
  \right)
  p_n(1-p_n)
\end{align*}
(we used (\ref{eq:expectation}) in the last equality).
We can rewrite this as
\begin{multline*}
  \K_N-\K_0
  =
  \frac12
  \left\|
    \sum_{n=1}^N
    (y_n-p_n)
    \Phi(p_n,x_n)
  \right\|^2_{\HHH}
  -
  \frac12
  \sum_{n=1}^N
  p_n(1-p_n)
  \left\|
    \Phi(p_n,x_n)
  \right\|^2_{\HHH},
\end{multline*}
which immediately implies~(\ref{eq:K29}).

\section{Forecast-continuity of feature mappings and kernels}
\label{app:continuity}

In this appendix we will prove,
essentially following \cite{steinwart:2001}, Lemma 3,
that the forecast-continuity of a kernel $\KKK$
on $[0,1]\times\mathbf{X}$
is equivalent to the continuity in $p$ of a feature mapping $\Phi(p,x)$
satisfying (\ref{eq:K1}).
As a byproduct,
we will also see that the forecast-continuity of a kernel $\KKK$
on $[0,1]\times\mathbf{X}$
can be equivalently defined by requiring that
\begin{itemize}
\item
  $\KKK((p,x),(p',x))$ should be continuous in $p$,
  for all $x\in\mathbf{X}$ and all $p'\in[0,1]$,
\item
  and $\KKK((p,x),(p,x))$ should be continuous in $p$,
  for all $x\in\mathbf{X}$.
\end{itemize}

In one direction the statement is obvious:
if $\Phi(p,x)$ is continuous in $p$,
the continuity of the operation of taking the inner product
immediately implies that $\KKK$ is forecast-continuous,
in both senses.

Now suppose that $\KKK$ is forecast-continuous,
as defined in the first paragraph of this appendix
(this is the apparently weaker sense of forecast-continuity).
To complete the proof,
notice that
\begin{align*}
  &\left\|
    \Phi(p,x) - \Phi(p_n,x)
  \right\|_{\HHH}\\
  &=
  \sqrt
  {
    \KKK((p,x),(p,x))
    -
    2\KKK((p,x),(p_n,x))
    +
    \KKK((p_n,x),(p_n,x))
  }\\
  &\to
  \sqrt
  {
    \KKK((p,x),(p,x))
    -
    2\KKK((p,x),(p,x))
    +
    \KKK((p,x),(p,x))
  }
  =
  0
\end{align*}
when $p_n\to p$ ($n\to\infty$).

\ifTR
\section{Derivation of the kernel of the Fermi--Sobolev space}
\label{app:kernel}

We first describe the standard reduction
of the problem of finding the kernel of an RKHS
to a variational problem.
Let $\KKK$ be the kernel of an RKHS $\FFF$ on $Z$.

Let $c\in Z$.
According to \cite{meschkowski:1962} (Satz III.3),
the minimum of $\left\|f\right\|_{\FFF}$ among the functions $f\in\FFF$
satisfying $f(c)=1$
is attained by the function $\KKK(\cdot,c)/\KKK(c,c)$.
Therefore, we obtain a function $k(\cdot,c)$ proportional to $\KKK(\cdot,c)$
by solving the optimization problem $\left\|f\right\|_{\FFF}\to\min$
under the constraint $f(c)=1$
(or under the constraint $f(c)=d$, where $d$ is any other constant).
It remains to find
the coefficient of proportionality
in terms of $k(\cdot,c)$.
If $\KKK(\cdot,\cdot)=\alpha k(\cdot,\cdot)$,
we have:
\begin{align*}
  \KKK(c,c)
  &=
  \left\|
    \KKK(\cdot,c)
  \right\|_{\FFF}^2;\\
  \alpha k(c,c)
  &=
  \alpha^2
  \left\|
    k(\cdot,c)
  \right\|_{\FFF}^2;\\
  \alpha
  &=
  \frac{k(c,c)}
  {
    \left\|
      k(\cdot,c)
    \right\|_{\FFF}^2
  }.
\end{align*}
Therefore,
the recipe for finding $\KKK$ is:
for each $c\in Z$
solve the optimization problem $\left\|f\right\|_{\FFF}\to\min$
under the constraint $f(c)=1$
(the completeness of RKHS implies that the minimum is attained)
and set
\begin{equation}\label{eq:KKK}
  \KKK(z,c)
  :=
  \frac
  {
    k(z,c) k(c,c)
  }
  {
    \left\|
      k(\cdot,c)
    \right\|_{\FFF}^2
  },
\end{equation}
where $k(\cdot,c)$ is the solution.

Now let us apply this technique to finding the kernel
corresponding to the Fermi--Sobolev space on $[0,1]$
with the norm given by (\ref{eq:norm}).
Let $c\in[0,1]$
and let $f$ be the solution
to the optimization problem $\left\|f\right\|_{\FFF}\to\min$
under the constraint $f(c)=1$
(because of the convexity of the set $\{f\in\FFF\st f(c)=1\}$,
there is only one solution).
First we show that the derivative $f'$ is a linear function on $[0,c]$ and on $[c,1]$,
arguing indirectly.
Suppose, for concreteness,
that $f'$ is not linear on the interval $(0,c)$;
in particular this interval is non-empty.
There are three points $0<t_1<t_2<t_3<c$ such that
\begin{equation}\label{eq:assume}
  f'(t_2)
  \ne
  \frac{t_3-t_2}{t_3-t_1}
  f'(t_1)
  +
  \frac{t_2-t_1}{t_3-t_1}
  f'(t_3).
\end{equation}
For a small constant $\epsilon>0$
(in particular, we assume
$
  2\epsilon<\min(t_1,t_2-t_1,t_3-t_2,c-t_3)
$),
let $g:[0,1]\to\bbbr$ be a smooth function such that
$\int_0^1 g(t)\D t=0$ and:
\begin{itemize}
\item
  $g(t)=0$ for $t<t_1-\epsilon$;
\item
  $g(t)$ is increasing for $t_1-\epsilon<t<t_1+\epsilon$;
\item
  $g(t)=t_3-t_2$ for $t_1+\epsilon<t<t_2-\epsilon$;
\item
  $g(t)$ is decreasing for $t_2-\epsilon<t<t_2+\epsilon$;
\item
  $g(t)=-(t_2-t_1)$ for $t_2+\epsilon<t<t_3-\epsilon$;
\item
  $g(t)$ is increasing for $t_3-\epsilon<t<t_3+\epsilon$;
\item
  $g(t)=0$ for $t>t_3+\epsilon$.
\end{itemize}
Since, for any $\delta\in\bbbr$
(we are interested in nonzero $\delta$ small in absolute value),
\begin{equation*}
  \left\|
    f+\delta g
  \right\|_{FS}^2
  =
  \left\|
    f
  \right\|_{FS}^2
  +
  2\delta
  \int_0^1
    f'(t) g'(t)
  \D t
  +
  \delta^2
  \int_0^1
    (g'(t))^2
  \D t,
\end{equation*}
the definition of $f$ implies
\begin{equation*}
  \int_0^1
    f'(t) g'(t)
  \D t
  =
  0.
\end{equation*}
However, as $\epsilon\to0$, the last integral tends to
\begin{equation*}
  f'(t_1)
  (t_3-t_2)
  -
  f'(t_2)
  (t_3-t_1)
  +
  f'(t_3)
  (t_2-t_1),
\end{equation*}
which cannot, by (\ref{eq:assume}),
be zero.

Once we know that $f$ is a quadratic polynomial to the left and to the right of $c$,
we can easily find
(this can be done conveniently using a computer algebra system)
that,
ignoring a multiplicative constant,
\begin{equation*}
  f(t)
  =
  3t^2+3c^2-6c+8
  =
  3t^2+3(1-c)^2+5
\end{equation*}
to the left of $c$
and
\begin{equation*}
  f(t)
  =
  3t^2+3c^2-6t+8
  =
  3(1-t)^2+3c^2+5
\end{equation*}
to the right of $c$.
By (\ref{eq:KKK}),
we can now find
\begin{equation*}
  \KKK(t,c)
  =
  \frac
  {
    f(t) f(c)
  }
  {
    \left\|
      f
    \right\|_{\FFF}^2
  }
  =
  f(t)/6,
\end{equation*}
which agrees with (\ref{eq:final}).
\fi
\ifWP\DFlastpage\fi
\end{document}